\documentclass[a4paper,fleqn]{cas-dc}

\usepackage{subfigure}
\usepackage[numbers]{natbib}
\usepackage{soul}
\usepackage{color}
\usepackage{float}
\usepackage{lineno}

\def\tsc#1{\csdef{#1}{\textsc{\lowercase{#1}}\xspace}}
\tsc{WGM}
\tsc{QE}

\begin{document}
\let\WriteBookmarks\relax
\def\floatpagepagefraction{1}
\def\textpagefraction{.001}

\shorttitle{}    


\title [mode = title]{Voice-Driven Mortality Prediction in Hospitalized Heart Failure Patients: A Machine Learning Approach Enhanced with Diagnostic Biomarkers}  



%

\author[1]{Mehmet Ali Sarsil}[orcid=0009-0000-0991-0578]
\fnmark[1]

\affiliation[1]{organization={Electronics and Communication Engineering Department, Istanbul Technical University},
            addressline={Maslak}, 
            city={Istanbul}, 
            country={Turkey}}
\affiliation[2]{organization={Department of Cardiology, Faculty of Medicine, Koç University},
            addressline={Zeytinburnu}, 
            city={Istanbul}, 
            country={Turkey}}

\affiliation[3]{organization={Department of Cardiology, School of Medicine, Kocaeli University},
            addressline={Izmit}, 
            city={Kocaeli}, 
            country={Turkey}}

\author[1]{Nihat Ahmadli}[orcid=0009-0005-2008-4533]
\fnmark[1]

\author[2]{Berk Mizrak}[orcid=0000-0001-7995-4488]
\author[3]{Kurtulus Karauzum}
\author[1]{Ata Shaker}
\author[2]{Erol Tulumen}
\author[3]{Didar Mirzamidinov}
\author[2]{Dilek Ural}[orcid=0000-0003-0224-1433]

\author[1]{Onur Ergen}[orcid=0000-0001-7226-4898]
\ead{oergen@itu.edu.tr}
\cormark[1]
\cortext[1]{Corresponding author. Department of Electronics and Communication Engineering, Istanbul Technical University Faculty of Electrical and Electronics, Sariyer, Istanbul, 34469, Turkey}

\fntext[1]{These three authors contributed equally to the study}


\begin{abstract}
Addressing heart failure (HF) as a prevalent global health concern poses difficulties in implementing innovative approaches for enhanced patient care. Predicting mortality rates in HF patients, in particular, is difficult yet critical, necessitating individualized care, proactive management, and enabling educated decision-making to enhance outcomes. Recently, the significance of voice biomarkers coupled with Machine Learning (ML) has surged, demonstrating remarkable efficacy, particularly in predicting heart failure. The synergy of voice analysis and ML algorithms provides a non-invasive and easily accessible means to evaluate patients' health.  However, there is a lack of voice biomarkers for predicting mortality rates among heart failure patients with standardized speech protocols. Here, we demonstrate a powerful and effective ML model for predicting mortality rates in hospitalized HF patients through the utilization of voice biomarkers. By seamlessly integrating voice biomarkers into routine patient monitoring, this strategy has the potential to improve patient outcomes, optimize resource allocation, and advance patient-centered HF management. In this study, a Machine Learning system, specifically a logistic regression model, is trained to predict patients' 5-year mortality rates using their speech as input. The model performs admirably and consistently, as demonstrated by cross-validation and statistical approaches (p-value < 0.001). Furthermore, integrating NT-proBNP, a diagnostic biomarker in HF, improves the model's predictive accuracy substantially.
\end{abstract}




\maketitle
\section{Introduction}\label{intro}

Heart Failure (HF) is a condition characterized by the heart's difficulty in pumping enough blood to meet the body's metabolic needs, affecting over 64.3 million individuals worldwide, and is on a critical rise \cite{general1, general2, general3, general4}. Between 2015 and 2018, approximately 6 million adults in the United States received a diagnosis of HF, with 13.4 \% succumbing to the condition, underscoring its profound impact on health \cite{burden1}. Moreover, heart failure not only affects patient health, morbidity, and mortality but also imposes a significant burden on healthcare costs and expenditures. For example, in the United States, the overall expenditure on HF care reached \$43.6 billion in 2020, with hospital admissions constituting 60\% of this total \cite{burden2, burden3, burden4, burden6}. At this point, it is imperative to understand and predict HF patient mortality rates to enable healthcare professionals to appropriately allocate resources, ensuring that individuals at higher risk of mortality receive treatment \cite{burden5}. In essence, this approach renders healthcare systems more effective, enabling patients to receive the necessary targeted care with greater ease.

In recent years, the synergy of voice analysis and Artificial Intelligence (AI) has emerged as a highly effective non-invasive method for diagnosing a wide spectrum of diseases, including heart failure. This innovative approach utilizes AI to discern specific voice changes associated with acute decompensated heart failure \cite{ai_voice0, ai_voice1, ai_voice2, ai_voice3}. The authors in \cite{ai_voice0}, for example, propose an Adaptive Intelligent Polar Bear Optimization-Quantized Contempo Neural Network (QCNN) model for diagnosing Parkinson's Disease using speech datasets. The process involves preprocessing speech signals, extracting features with Statistical Time Frequency Renyi, optimizing relevant features with the AIPB algorithm, and classifying the speech signals using the QCNN technique. Another study in \cite{ai_voice1} presents the development of two artificial neural networks (ANNs) designed to recognize vocal distortions caused by heart failure (HF) in individuals. The researchers collected voices from 142 individuals and applied various signal analysis techniques such as statistical analysis, FFT, discrete wavelet transform, and Mel-Cepstral analysis to extract features. The resulting ANNs achieved very high performance, demonstrating that voice analysis can significantly enhance HF recognition and early treatment. These studies along with others solidify the significant potential of voice biomarkers in the diagnosis of Heart Failure disease\cite{vocalbiomarker1, vocalbiomarker2, vocalbiomarker3}.

Machine Learning (ML) and Artificial Intelligence algorithms have also demonstrated efficacy in the analysis of mortality among patients diagnosed with heart failure \cite{ml_mortality1, ml_mortality2, ml_mortality3, ml_mortality4, ml_mortality5, ml_mortality6}. To illustrate, \cite{ml_mortality1} developed an AI-based algorithm for predicting in-hospital mortality and long-term mortality in acute heart failure (AHF) patients, using demographic information, electrocardiography, echocardiography, and laboratory data. Furthermore, Angraal et al. proposed the use of machine learning methods to develop and validate risk assessment models for predicting mortality and hospitalization in HFpEF (Heart Failure with preserved Ejection Fraction) patients, using medical data from the TOPCAT trial \cite{ml_mortality3}. Another study in \cite{ml_mortality4} put forth a Deep Learning System based on a Multi-head Self-attention Mechanism (DLS-MSM) for predicting mortality in heart failure (HF) patients. Addressing issues like missing values and data imbalance, the system employs an indicator vector, convolutional neural network (CNN) with different kernel sizes, multi-head self-attention, and focal loss function. 

Nevertheless, the aforementioned models underwent training using health-related medical data, rendering these methodologies notably invasive and inefficient. It is also noteworthy that the integration of voice analysis, an emerging non-invasive diagnostic tool, with AI for mortality prediction in HF remains an underexplored frontier. Despite the well-established success of the previously mentioned AI models in mortality prediction within the HF domain, a noticeable gap exists in the literature, as no studies have connected ML and voice analysis to predict mortality rates in patients with HF using standardized speech protocols. Thus, in this paper, we present an important methodology aimed at bridging this crucial gap, at the convergence of AI and voice analysis. We developed an ML-based digital voice monitoring tool tailored to predict 5-year mortality rates among heart failure patients. Specifically, the logistic regression with a Ridge regularization was used for the training of our machine learning model. It yields an Acoustic Predictor, which comprises crucial vocal features and exhibits a robust correlation with the 5-year mortality rate (p < 0.001). The model's exceptional performance, validated through cross-validation and statistical methods (p-value <0.001), establishes it as a groundbreaking advancement in the integration of voice analysis and AI for personalized healthcare. Following a comprehensive statistical evaluation of additional features, the model underwent retraining with the integration of a diagnostic biomarker for heart failure: NT-proBNP. This refinement significantly enhanced the predictive accuracy of the model, underscoring the pivotal role of NT-proBNP in mortality rate predictions for patients with HF. Two additional conventional machine learning methods, specifically Decision Tree (DT) and Random Forest (RF), were further employed and their results are presented. Finally, a number of parameters that were fed to the algorithm are elucidated for clinical interpretability.

\subsection{{Major Contributions:}}
\begin{enumerate}
    \item {Novel Integration of Voice Analysis and Machine Learning for Mortality Prediction:} We introduce an innovative ML-based approach that leverages voice biomarkers to predict 5-year mortality rates in heart failure patients. This is the first study to utilize voice analysis combined with machine learning for this purpose, filling a significant gap in the literature.
    \item Development of a Non-Invasive Predictive Tool: Our approach employs logistic regression with Ridge regularization to create an Acoustic Predictor. This tool identifies key vocal features that are strongly correlated with the 5-year mortality rate (p < 0.001). The non-invasive nature of voice analysis makes this tool highly practical and accessible for continuous patient monitoring.
    \item Enhanced Predictive Accuracy with NT-proBNP Integration: By incorporating the diagnostic biomarker NT-proBNP into our model, we significantly improved its predictive accuracy. This integration highlights the critical role of NT-proBNP in mortality prediction and underscores the enhanced performance of our model compared to those relying solely on traditional clinical data.
\end{enumerate}

\subsection{{Structure of the Manuscript:}}
\begin{enumerate}
    \item \textbf{Introduction:} Background on heart failure, the potential of voice biomarkers, the gap in existing research, and our approach.
    \item \textbf{Theoretical Background:} Theoretical relationship among vocal features, heart failure, and mortality rate is explored.
    \item \textbf{Data Collection:} Description of the study population, data collection methodology, and voice recording protocol.
    \item \textbf{Data Preprocessing:} Description of feature selection and engineering process.
    \item \textbf{Model Development:} Detailed explanation of the machine learning model and integration of NT-proBNP.
    \item \textbf{Results:} Presentation of the statistical analysis, model's performance, and comparison with other machine learning methods.
    \item \textbf{Discussion and Future Work:} Interpretation of findings, limitations and future prospective
    \item \textbf{Conclusion:} Summary of key findings and contributions
\end{enumerate}

\section{Theoretical Background}

Voice features provide a unique window into the physiological and pathological state of the cardiovascular system, particularly in the context of heart failure (HF). HF is characterized by reduced cardiac output, leading to systemic effects such as fluid retention, pulmonary congestion, and decreased oxygen delivery to tissues. These systemic effects can manifest in the vocal apparatus, resulting in distinct changes in voice characteristics. For instance, studies have demonstrated that an increased pause ratio in speech, a reflection of respiratory distress and fatigue, correlates strongly with elevated NT-proBNP levels, a biomarker of HF severity \cite{vocalbiomarker1}. This connection is grounded in the physiological interplay between cardiac and respiratory functions, wherein compromised cardiac output directly impacts respiratory efficiency and, consequently, voice production. Thus, voice analysis offers a non-invasive, real-time indicator of underlying HF pathology and patient prognosis.

Additionally, the predictive power of voice features extends to their significant associations with coronary artery disease (CAD), a condition often co-morbid with HF. Research has shown that specific voice signal characteristics, such as frequency and amplitude variations, are independently associated with CAD, even when traditional risk factors are accounted for \cite{theory1}. This robust association underscores the physiological basis for using voice analysis as a diagnostic and prognostic tool in cardiovascular diseases. 

The vagus nerve, which is integral to voice production, also plays a crucial role in the autonomic control of the heart, influencing heart rate and variability. Dysfunctions in the vagus nerve and other cranial nerves involved in voice production can thus reflect underlying cardiovascular abnormalities, including HF \cite{MITTAPALLE202235, maor2020}. The association between vocal biomarkers and adverse outcomes in HF patients, such as hospitalization and mortality, suggests that these voice features can serve as non-invasive indicators of disease severity and progression. This connection is supported by findings that demonstrate significant correlations between voice signal characteristics and increased risk of death in HF patients, independent of traditional risk \cite{maor2020}.











\begin{figure*}[t]
	\centering
		\includegraphics[width=1.1\linewidth, height=0.6\linewidth]{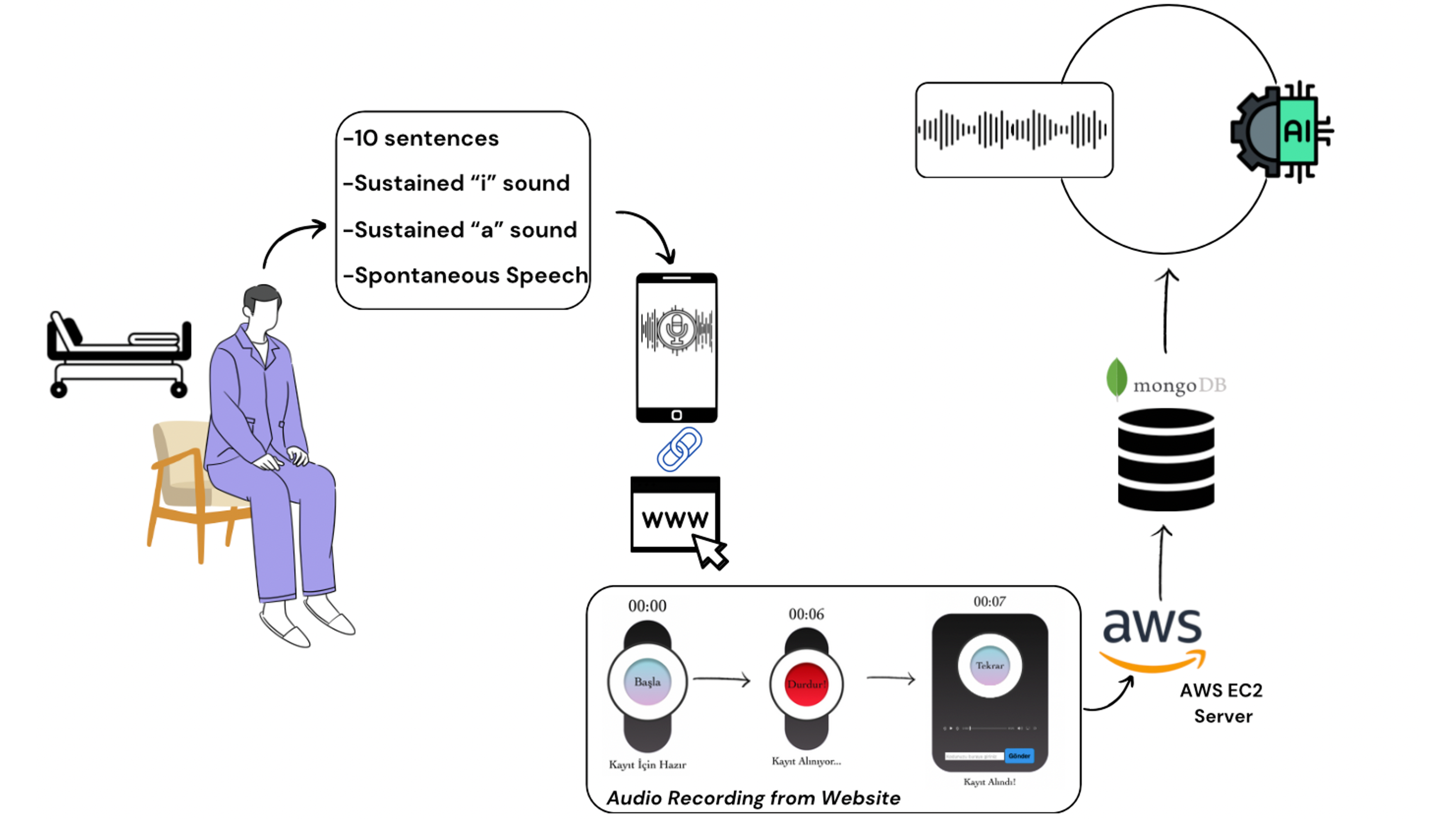}
	  \caption{Collection of patient voice illustrated}\label{fig: website}
\end{figure*}

\section{Data Collection}\label{sec: DataCollection}

\subsection{Study Population} \label{sec: population}

The study involved the inclusion of 29 patients admitted to the hospital for decompensated acute heart failure (HF) between June and September 2023. To be eligible, participants were required to be over 18 years of age and have a history of HF. All patients meeting these criteria were invited to participate in the study before their discharge within the specified timeframe. The cohort comprised 26 males and 3 females. The observed gender distribution among patients may be indicative of the inherent demographic characteristics of individuals seeking medical attention for decompensated acute HF \cite{gender}. All patients in the cohort had pre-existing chronic heart failure. The research specifically concentrated on native Turkish-speaking patients, chosen for their linguistic diversity, which facilitates algorithm generalization across various language types. Targeting individuals aged 18 or older with acute decompensated heart failure and confirmed reduced left ventricular ejection fraction (<40\%), the research aimed to achieve a comprehensive representation of this specific population.

Exclusion criteria were established to omit children under 18 years old and individuals with conditions such as mildly reduced or preserved ejection fraction (>40\%), acute coronary syndrome, significant respiratory distress affecting text readability, malignancies impacting voice quality, congenital heart disease, mechanical heart valves, reduced glomerular filtration rate (<25 mL/min), active respiratory infections, moderate or severe chronic obstructive pulmonary disease, a history of laryngeal or oropharyngeal surgeries, fever, or recent vaccination (within the past week).

Furthermore, the study excluded those unable to provide informed consent, illiterate individuals, and patients with articulation disorders. Table \ref{Table: patients}. provides detailed baseline characteristics of the study cohort. Ethical approval was obtained from the human research ethics committee, and written informed consent was secured from each patient before inclusion in the study.

\subsection{Website Development}\label{sec: website}

A dedicated website, designed for audio recording at a sampling rate of 44.1 kHz and 16 bits per sample, was developed to streamline the process of collecting data from patients. This website is hosted on an EC2 instance server via Amazon Web Services (AWS), and the recorded voice data is methodically stored in a MongoDB database. A detailed depiction of the process for collecting patient voice information is presented in Fig. \ref{fig: website}. 

\begin{figure*}[t]
	\centering
		\includegraphics[width=1\linewidth, height=0.69\linewidth]{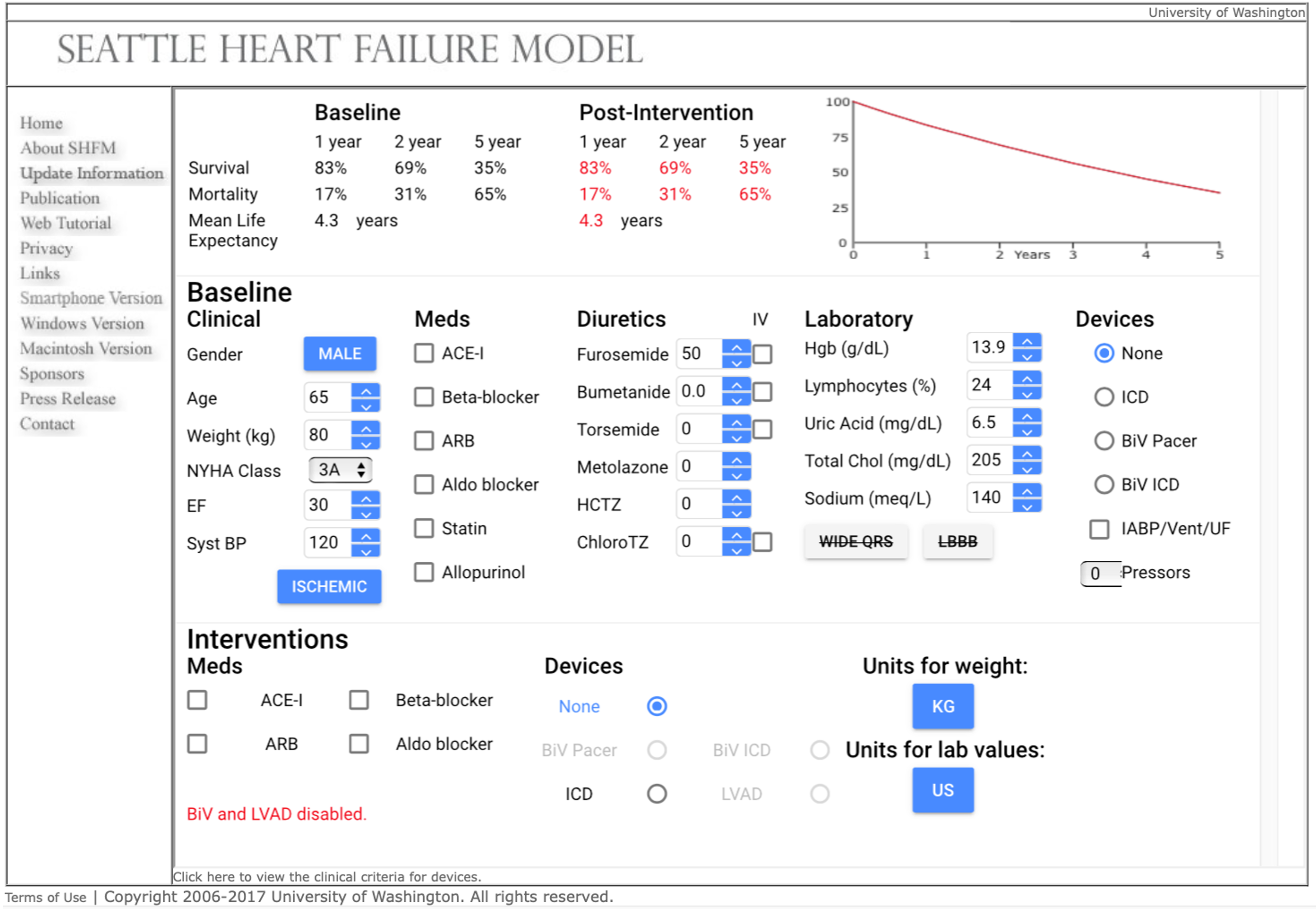}
	  \caption{Interface of a website for SHFM}\label{fig: shfm}
\end{figure*}

\subsection{CAPE-V Protocol and Voice Recordings} \label{sec: capev}

The CAPE-V (Consensus Auditory-Perceptual Evaluation of Voice) was developed as a clinical tool for auditory-perceptual voice assessment \cite{capev}. Its primary aim is to characterize the auditory-perceptual features associated with a voice issue, facilitating effective communication among healthcare professionals. The protocol was formulated by the American Speech-Language-Hearing Association’s (ASHA) Special Interest Division 3, Voice and Voice Disorders, and achieved consensus adoption at a conference held in 2002 at the University of Pittsburgh \cite{capev}.

The assessment involves gathering voice samples from patients across four sections:

\begin{itemize}
\item Six sentences specifically crafted based on different phonetic contexts.
\item Sustained vowel "a" for 5 seconds.
\item Sustained vowel "i" for 5 seconds.
\item Conversational speech, including responses to at least two of three daily questions.
\end{itemize}

Patients are instructed to record their voices using the designated website, following the Turkish version of the CAPE-V protocol \cite{capev_turk}, which includes the four sequential segments mentioned above. The average duration of each segment collected is 35.4, 3.4, 2.7, and 47.5 seconds, respectively. Voice recordings were consistently captured using a single iPhone 14 Pro Max cell phone by cardiology specialists at Kocaeli University Hospital. To maintain uniformity, voice recordings were conducted on a specially developed website mentioned in Section \ref{sec: website} (recordituku.com.tr). To ensure standardized and clear recordings, the cell phone was positioned 20-30 cm from the patient's mouth, and recordings were made in a quiet environment to minimize background interference.

\subsection{Seattle Heart Failure Model: Target Label} \label{SHFM}

In developing the machine learning-based model to predict the 5-year mortality rate through patient voice analysis, we employed the Seattle Heart Failure Model (SHFM) as the ground truth target label for our dataset. The SHFM is a risk prediction tool used in the field of cardiology to estimate the prognosis of patients with heart failure \cite{shfm}. Designed to aid clinicians in assessing expected survival and risk of adverse events in heart failure individuals, SHFM offers user-friendly online calculators and software tools, as illustrated in Fig. \ref{fig: shfm}.

SHFM provides four predictive parameters estimating survival rates in heart failure patients, with "SHFM 5" specifically indicating the 5-year mortality rate. Expressed as a percentage, this parameter represents the probability of a patient deceasing within 5 years and is selected as the target variable in our classification problem. To facilitate binary classification, a predetermined 5-year mortality threshold of 50\% was established. Patients with a probability exceeding 50\% of succumbing to mortality within 5 years (according to SHFM) were assigned a prognostic label of 1, while those falling below this threshold were assigned 0.

The establishment of a 50\% 5-year mortality risk threshold in chronic heart failure (HF) is well-grounded, considering epidemiological data that illustrates a notable decline in survival rates from 80-90\% at 1 year to 50-60\% at 5 years, indicative of a substantial rise in mortality risk during this interval. This is supported by population-level epidemiological evidence demonstrating mean survival rates in chronic HF of 80–90\%, 50–60\%, and 30\% at 1, 5, and 10 years respectively. Consequently, the ability to anticipate such heightened mortality risk over a 5-year period can aid clinicians in tailoring treatments more effectively based on a designated risk threshold \cite{fifeyear1, fifeyear2, fifeyear3, fifeyear4, fifeyear5}.

Additionally, this resulted in a nearly equal distribution of binary target labels as Table {\ref{Table: patients}} illustrates, effectively mitigating any potential data imbalance concerns.

\begin{table*}[h] 
\caption{\label{Table: disvoice_param} Vocal parameters and their meanings}
    \centering
    \begin{tabular}{| p{1.5cm} | p{3.5cm} | p{1.2cm} | p{3.2cm} | p{1.8cm} | p{3.5cm} |}
        \hline
        \multicolumn{2}{|c|}{Glottal Features}
        & \multicolumn{2}{c|}{Phonation Features}
        & \multicolumn{2}{c|}{Prosody Features} \\
        \hline
        GCI & Variability of time between consecutive glottal closure instants
        & F'0 & The rate of pitch change in a voice signal
        & Tilt of linear estimation of F0 for each voiced segment
        & The slope or incline in the pitch or fundamental frequency across that segment \\
        \hline
        Average of OQ & Average opening quotient (OQ) for consecutive glottal cycles -> rate of opening phase duration/duration of the glottal cycle
        & F''0 & The rate of change in the rate of pitch change in a voice signal & Energy on first segment & The initial loudness of a voiced part of speech \\
        [15.5ex]\hline
        Variability of OQ & Variability of opening quotient (OQ) for consecutive glottal cycles -> rate of opening phase duration /duration of glottal cycle & Jitter & The irregularity or variability in the time intervals between consecutive pitch cycles in a voice signal & Energy on last segment & The loudness of the final portion of a voiced speech segment \\
        [16.5ex]\hline
        Average of NAQ	& Average Normalized Amplitude Quotient (NAQ) for consecutive glottal cycles -> ratio of the amplitude quotient and the duration of the glottal cycle & Shimmer & The variations in the amplitude or intensity of consecutive pitch cycles in a voice signal & Voiced rate & Number of voiced segments per second \\
        [15.5ex]\hline
    \end{tabular}
\end{table*}

\section{Data Preprocessing}\label{sect: preprocessing}

\subsection{Feature Extraction} \label{sect: FeatureExtraction}

We employed the ‘Disvoice’ library in Python programming language as a crucial component of our data processing pipeline to extract relevant acoustic features from patients' voices. This framework computes glottal, phonation, and prosody features from raw speech files \cite{disvoice1, disvoice2, disvoice3, disvoice4}. 

For instance, the calculation of glottal features involves computing descriptors derived from glottal source reconstruction. In the case of Disvoice, the focus is on evaluating nine descriptors, including the "Average Harmonic Richness Factor (HRF)." This descriptor calculates the average ratio between the sum of the amplitudes of harmonics and the amplitude of the fundamental frequency, offering insights into the richness of harmonic content in the voice \cite{disvoice1}. The phonation features computed by the model reveal abnormal and impaired patterns in vocal fold vibration, assessed in terms of stability measures. Disvoice predominantly concentrates on evaluating seven descriptors, encompassing parameters such as Jitter and Shimmer. Jitter quantifies variations in the fundamental period of the voice signal, while Shimmer assesses variations in voice signal amplitude \cite{disvoice3}. Prosody features encapsulate aspects of speech conveying information about timing, pitch variations, and volume levels in natural speech. Traditionally assessed using metrics derived from F0 and energy contours, Disvoice emphasizes the computation of 103 descriptors, including the F0-contour parameter capturing pitch fluctuations over time, providing insights into the rhythmic, intonational, and emotional aspects of speech \cite{disvoice4}. 

Vocal features are computed on frames obtained by dividing the audio signals. Additionally, calculations of statistical parameters— average, standard deviation, maximum, minimum, skewness, and kurtosis — over the vocal features across the frames were included in the extracted features. Table \ref{Table: disvoice_param}. provides examples of vocal parameters derivable from the three feature sets.

In our feature extraction process, we exclusively derived glottal features from sustained vowels, a choice made owing to the distinctive characteristics of sustained vowels. Sustained vowels provide a controlled and stable environment, facilitating a focused study of glottal behavior \cite{glottal}. Phonation features, on the other hand, were extracted from all segmented voice sections, while prosody features were extracted from sustained vowels and the CAPE-V protocol sentences mentioned in Section \ref{sec: capev}. The prosody features were not extracted from conversational speech segments. This exclusion was deliberate as conversational speech might include spontaneous pauses related to the thought process, potentially introducing variability in prosody features. Given that prosody features involve measures derived from speech duration analysis, including the length of pauses, extracting them from conversational speech could yield misleading results. Therefore, to ensure the reliability of prosody features, we limited extraction to sustained vowels and carefully chosen sentences from the CAPE-V protocol.

\subsection{Feature Engineering} \label{sec: feature engineering}

A total of 494 features were extracted from each patient across all speech segments using the aforementioned Disvoice library.

\begin{figure*}[t] 
	\centering
		\includegraphics[width=1.0\linewidth, height=0.43\linewidth]{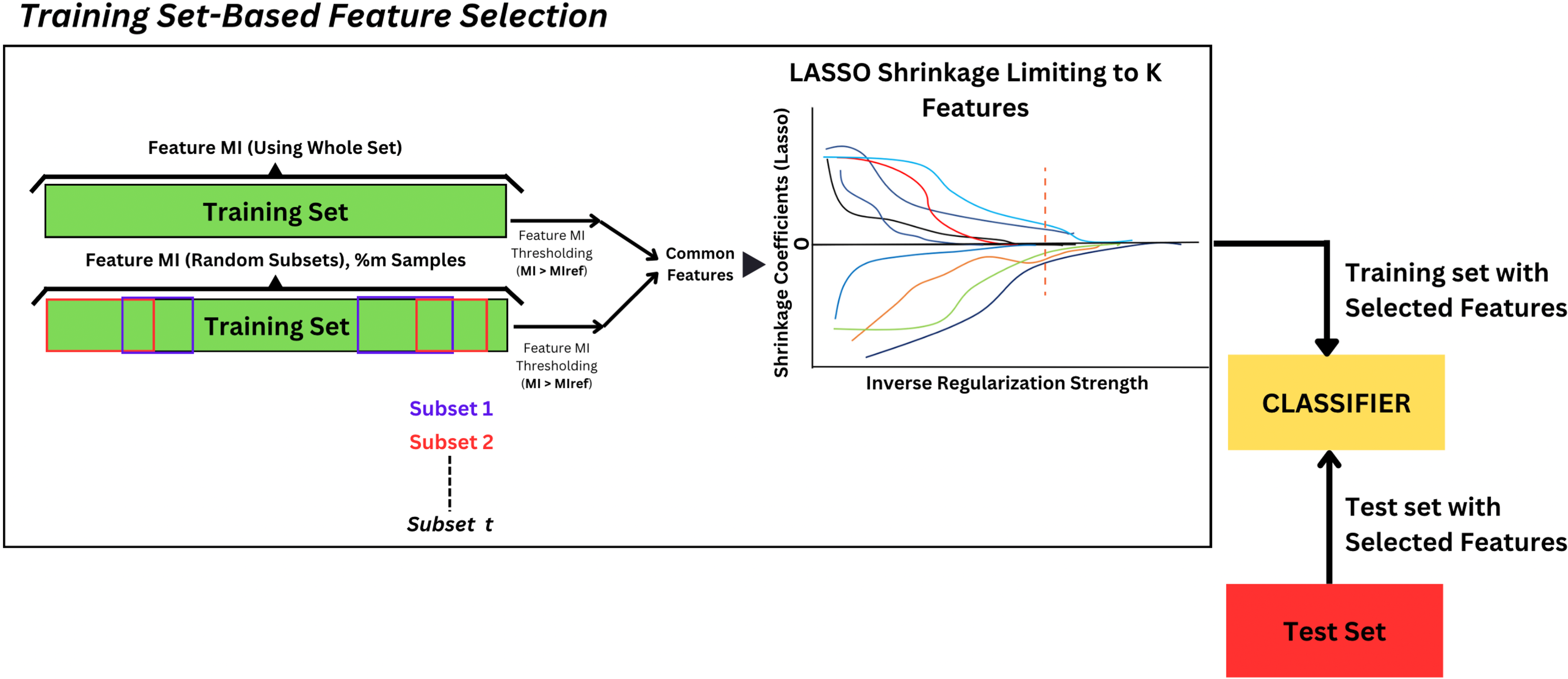}
    
	  \caption{\label{fig:preprocessing}The multi-step feature preprocessing pipeline is meticulously crafted to curate the most representative features for predicting the target label in each classification model employed in this study. Initially, the pipeline assigns reliable Mutual Information (MI) scores to each feature in the training set, followed by a subsequent step involving a backward elimination algorithm coupled with the LASSO shrinkage technique.}
\end{figure*}

The preprocessing steps were meticulously designed and empirically refined using the training set to select the most representative features for predicting the target label, addressing several issues with our feature set and enhancing the effectiveness of the machine learning model. Firstly, a high-dimensional dataset consisting of 494 features is susceptible to "the curse of dimensionality" \cite{curse}. This phenomenon can render a machine learning model prone to overfitting, increase computational intensity, and necessitate larger datasets for effective generalization. Secondly, the presence of collinearity among most of the 494 features posed a risk of model instability, hindered interpretability, and increased the likelihood of overfitting, collectively impacting the accuracy and reliability of predictions \cite{overfitting}. To mitigate these issues, the initial step involves hard-thresholding features by removing them based on a predefined Mutual Information (MI) score. This score quantifies the extent of association between variables, indicating how much information on one variable can be inferred by observing another \cite{Latham2009}. Filtering based on an MI score threshold follows a two-step algorithm. First, the MI score of each feature is computed using the entire training set to identify features with high MI scores across the dataset. Subsequently, each feature's MI score is recalculated by averaging individual MI scores computed over randomly sampled subsets of the training set, each comprising 80\% of the original training set size, and totaling 15 subsets, these subsets are repeatedly shuffled before sampling to ensure robustness in the calculation. The features that consistently exceeded the predetermined MI score threshold of 0.105 in both evaluation steps, totaling 10 features, were retained for further preprocessing. This two-step approach aims to identify features that robustly align with the calculated MI scores while filtering out features that exhibit sporadic high MI scores based on randomly formed subsets. This initial hard-thresholding step is crucial for reducing the total number of features, which in our case is 494, to a manageable quantity that can be effectively utilized in more advanced algorithms capable of feature elimination while also considering inter-correlation among features and maximizing mutual information. The Least Absolute Shrinkage and Selection Operator (LASSO), utilizing L1 regularization, in conjunction with a backward feature elimination method, serves as the final feature selection algorithm for this purpose \cite{lassobackward1, lassobackward2, lassobackward3, Hastie2001}. A logistic regression estimator with LASSO regularization, utilizing a corresponding coefficient of 1.2, is trained on the training set and deployed through a Python 3.11 API sklearn.feature-selection.RFE(), of scikit-learn library \cite{scikit-learn}. In this configuration, LASSO shrinks the coefficients of features based on their explanatory power regarding the target label, while the recursive feature elimination method iteratively selects features by considering progressively smaller sets. Initially, the estimator is trained on the full feature set, and the importance of each feature is assessed using LASSO-assigned coefficients. Subsequently, the least important features are pruned from the current set, and this process is recursively repeated until the desired number of features, set to 5, is reached \cite{sklearn_api}. These preprocessing steps facilitated the creation of a more interpretable and robust model, allowing for meaningful insights and conclusions. Figure {\ref{fig:preprocessing}} provides a visual representation of these sequential steps.

\subsubsection{Description and Characteristics of Utilized Vocal Features}
\label{sec:featureExplanation}

\begin{figure*}[b] 
	\centering
		\includegraphics[width=1\linewidth, height=0.45\linewidth]{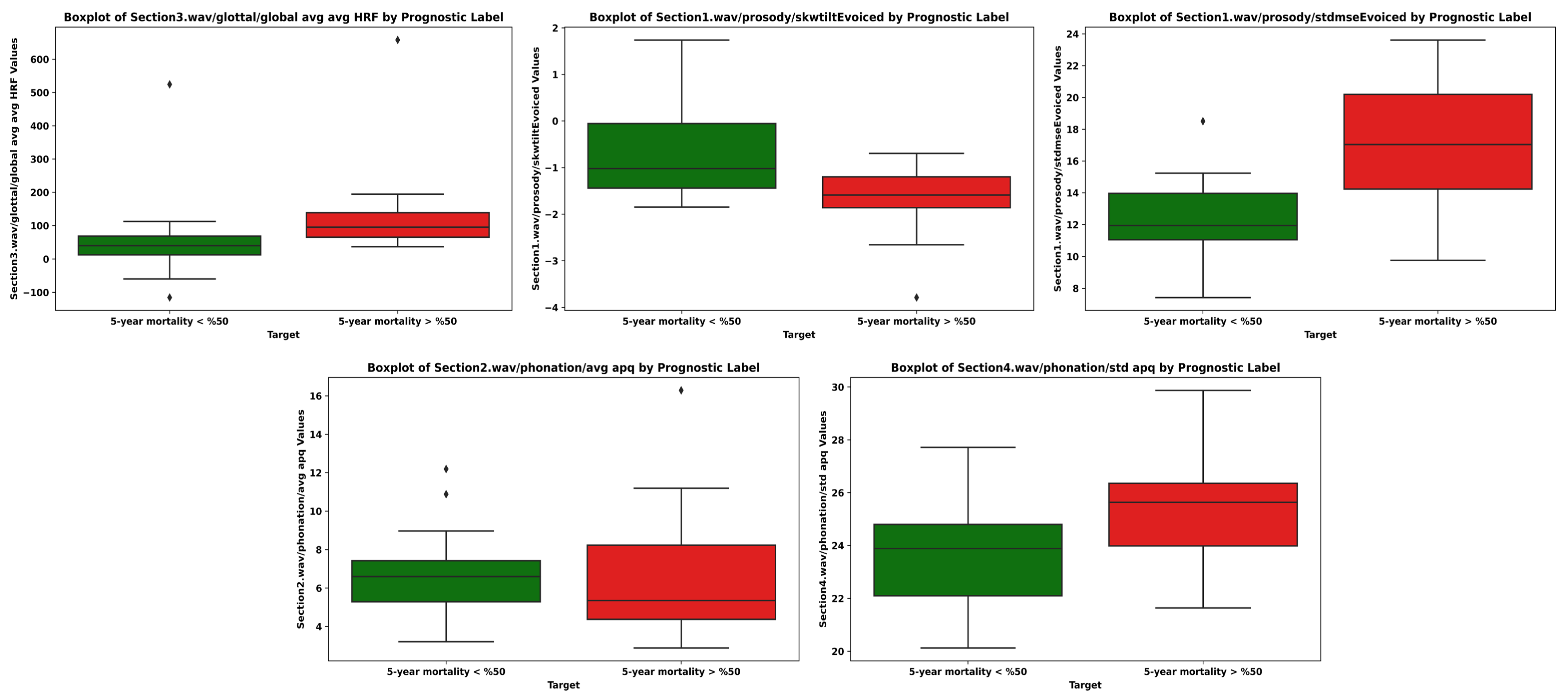	}
    
	  \caption{\label{fig:boxplot}Box plots illustrating the distributions of five features utilized in the modeling process, with respect to the prognosis target label}
\end{figure*}

Figure {\ref{fig:boxplot}} illustrates the box-plot distributions of five vocal features utilized in training the logistic regression model within the train-test split methodology. The feature names follow a convention where each component separated by a slash delineates a specific aspect of the feature. For instance, in the "Section2.wav/phonation/avg apq" feature, "Section2.wav" denotes a segment of the audio, specifically the sustained vowel "a" for 5 seconds, and also signifies the name of the data file. "Phonation" describes the category or aspect of the feature being analyzed, while "avg apq" specifies the specific measurement or statistic calculated within the phonation aspect, namely the average of the amplitude perturbation quotient (apq) values. Another example is 'Section3.wav/glottal/global avg avg HRF,' where the glottal feature represents the global harmonic richness factor, calculated as the ratio of the sum of the amplitude of the harmonics and the amplitude of the fundamental frequency, providing information from section 3, which corresponds to the sustained vowel "i" for 5 seconds. Additionally, 'Section1.wav/prosody/skwtiltEvoiced' signifies the statistical skewness parameter of spectral tilt, describing the distribution of energy across different frequency bands. 'Section1.wav/prosody/stdmseEvoiced' represents the statistical standard deviation of the mean squared error (MSE) between the original F0 values and the reconstructed linearly estimated F0 values for each voiced segment using linear regression, pertinent to section 1. Moreover, 'Section4.wav/phonation/std apq' denotes the standard deviation of the amplitude perturbation quotient (apq) values calculated over section 4, corresponding to conversational speech.

\section{Model Development}
Given the binary classification nature of our task, we opted for Logistic Regression (LR) as the appropriate modeling technique \cite{logistic}. In the realm of machine learning and statistics, LR is a predictive analytic method grounded in the probability concept. It aids decision-making in scenarios with two choices, such as diagnosing a medical condition or predicting a patient's risk. LR operates by analyzing data and discerning patterns leading to one of two outcomes, for example, determining if a patient has a particular disease (yes or no), or in our specific case, predicting whether a patient with heart failure will survive within 5 years (yes or no).

In LR, we start by creating a linear combination of the input features X and a set of coefficients ($\theta$) denoting the weights of each feature plus an intercept ($\theta_0$):

\begin{equation} \label{eq: z}
    z=\ \theta_0\ +\ \theta_1.x_1\ +\ \theta_2.x_2\ +\ \ \ldots.\ {\ +\ \ \theta}_n.x_n 
\end{equation}

where $z$ is the linear combination, $\theta_0$ is the intercept (also known as the bias term), and $\theta_i$ are the weights (coefficients) for the $i^th$ feature. This linear combination, $z$, represents a measure of evidence for the positive class (class 1).

Once fitted to a dataset, LR predicts the probability of a positive class $p(y_i=1 | X_i)$ with the help of a logistic function (also called sigmoid function):

\begin{equation} \label{eq: sigmoid}
\ p=\ \frac{1}{1+\ e^{-(\theta_0\ +\ \theta_1.x_1\ +\ \theta_2.x_2\ +\ \ \ldots.\ {\ +\ \ \theta}_n.x_n)}}\ 
\end{equation}

where e is the base of the natural logarithm and p is the probability of belonging to class 1.

Subsequently, to make a classification decision, a threshold (often set at 0.5) is applied to the predicted probability $p$. If $p$ is greater than or equal to this threshold, the data point is classified as belonging to class 1; otherwise, it is classified as class 0.

To effectively train the model, optimal values for the coefficients $\theta$ need to be determined. This is typically achieved through a process known as Maximum Likelihood Estimation \cite{mle}. The objective is to identify the values of $\theta$ that maximize the likelihood of the observed data given the model. Afterward, a cost or loss function \cite{loss} is employed to quantify the error between the predicted probabilities and the true class labels in the training data:

\begin{equation}
\begin{split}
\mathcal{L}(y_i, p_i) &= \frac{1}{m}\sum_{i=0}^{m} [y_i \cdot \log(p_i) \\
&\quad+ (1-y_i) \cdot \log(1-p_i)] 
\end{split}
\label{eq:loss}
\end{equation}

where $m$ is the number of training data samples, $y_i$, and $p_i$ are the actual class label (0 or 1) and predicted probability of class 1 for the $i^th$ sample, respectively.
 
The optimization algorithm then seeks to minimize this loss function by updating the coefficients $\theta$ in each iteration until the loss converges to a minimum value.

Regularization serves the purpose of preventing overfitting by introducing a penalty term to the model's cost function \cite{regularization}, which assesses the model's fit to the data. Binary class logistic regression with the addition of a regularization term aims to minimize the cost function in Eq. \ref{eq:loss}. Numerous regularization techniques with the penalization arguments, such as $l1$ and $l2$, exist. The penalization argument can be regarded as a hyperparameter of the LR model, a parameter that is used to configure an ML model \cite{hyperparameter}.

To implement the model on our data, the available data from 29 patients was initially split into training and test sets, with a test ratio of 0.35, facilitating model evaluation. This division resulted in 18 training samples and 11 test samples.

Subsequently, GridSearchCV \cite{gridsearch}, a hyperparameter tuning technique, was employed to systematically search and select the optimal hyperparameters associated with the building blocks of the preprocessing algorithm and final LR model. This involved an exhaustive search over a predefined hyperparameter grid, where various combinations of hyperparameters, such as MI score threshold, LASSO regularization parameter for the amount of shrinkage for the preprocessing blocks and regularization strength, and penalty type for the final LR model, were tested. The selection criteria were based on performance metrics like accuracy, precision, and recall, aiming to identify the hyperparameters that produced the best classification results. Following the hyperparameter tuning, the logistic regression model, equipped with its best-selected hyperparameters, was trained using the 18 training samples and evaluated with the 11 test samples. The optimal hyperparameters for LR, determined through GridSearchCV, along with their corresponding meanings, are presented in Table {\ref{Table: Grid}}. Moreover, the code for feature extraction, preprocessing, and modeling has been publicly shared on
\href{https://github.com/malisarsil/heartFailurePrognosisfromVoice}{Github(https://github.com/malisarsil/
heartFailurePrognosisfromVoice)}.

\begin{table*}[h]
\centering
    \caption{\label{Table: Grid} The optimal hyperparameters chosen by GridSearchCV}
    \small\addtolength{\tabcolsep}{2.5pt}
    \begin{tabular}{| p{3.5 cm} | p{3.5 cm} | p{8.5cm} |}
    \hline
     Hyperparameter & GridSearchCV output & Explanation \\
     \hline
     “C”	& 0.2 & The inverse of the regularization strength. A smaller "C" value results in stronger regularization, while a larger "C" value reduces the strength of regularization. \\
     [6.0ex]\hline
     “penalty” & “L2” & Works in conjunction with the "C" parameter to determine the type of regularization applied to the model. L2 regularization encourages the model to keep all features but shrink the coefficients toward zero. \\
     [9.0ex]\hline
     “solver” & “liblinear”	& Responsible for choosing the optimization algorithm used to fit the logistic regression model to the training data. "liblinear" is a solver, particularly for binary classification problems. It is known for its efficiency on small-sized datasets and is suitable for problems with both L1 and L2 regularization. \\ 
     [11.5ex]\hline
    \end{tabular}
\end{table*}

\textbf{Acoustic Predictor Definition} \label{sec: predictor}

For the statistical analysis, z in Eq. (\ref{eq: z}) is described as a unitless quantity called the “acoustic predictor”, which is a combination of five selected features, as mentioned in Section \ref{sec: feature engineering}. 

\section{Results} \label{sec: results}

\subsection{Statistical Analysis} \label{sec: statistics}
The Study population was categorized into two groups based on the SHFM 5-year mortality predictions. Patients with less than 50\% 5-year mortality prediction (n=15) were in the first group, whilst patients with more or equal to 50\% 5-year mortality prediction (n=14) were in the second group. Variables were expressed as median with interquartile range and as frequency (\%) for categorical variables. The two groups were compared using a t-test test for continuous variables, including the acoustic biomarker as an outcome of our model, and $\chi^2$ test for categorical variables. Statistical significance was accepted for a 2‐sided P<0.05. The statistical analyses were performed with SPSS software (Version 28.0.0.0, 2023). All supporting data are available within the article and its online supplementary file.

The final study population comprised 29 decompensated patients with pre-existing chronic heart failure. The baseline characteristics of the study cohort are given in Table \ref{Table: patients}. For simplicity, patients with less than 50\% SHFM 5-year mortality rate are described as patients with lower mortality, whilst patients with equal or more than 50\% SHFM 5-year mortality rate are described as patients with higher mortality. Patients with a lower mortality rate had an average age of 69.07 ± 4.80 years, while those with higher mortality averaged 66.86 ± 10.49 years. 100\% of the patients in the lower mortality group were male, compared to 78.6\% in the higher mortality group. Heart Failure Duration was notably longer in the higher mortality group, averaging 8.80 ± 3.96 years, compared to 5.08 ± 3.37 years in the lower mortality group. Significant disparities in NT-proBNP levels were observed, with the higher mortality group recording 19845.79 ± 15090.14 and the lower group registering 8119.67 ± 8190.55. Daily Furosemide Dose Requirement was higher in the higher mortality group at 387.69 ± 121.53, as opposed to 256.00 ± 159.50 in the lower mortality group. Both groups had high percentages of patients on beta-blockers, with 86.7\% in the lower mortality group and 85.7\% in the higher mortality group. Notably, there was a significant difference in the use of lipid-lowering drugs, with 80\% usage in the lower mortality group and only 28.6\% in the higher mortality group. NYHA classifications showed clear differences. In the lower mortality group, 86.7\% of patients were classified as NYHA III, whereas in the higher mortality group, 64.3\% were NYHA IV. Dyspnea was observed in 73.3\% of the lower mortality group and 85.7\% of the higher mortality group.  Pleural effusion was more prevalent in the higher mortality group, with 42.9\% of patients affected, compared to 20\% in the lower mortality group.

\begin{table*}[t]
\centering
    \caption{\label{Table: patients} Patient Characteristics and statistical test results}
    \small\addtolength{\tabcolsep}{2.5pt}
    \begin{tabular}{| p{6.0 cm} | p{4.0 cm} | p{4.0cm} | p{1.0cm} |}
    \hline
     & 5-year mortality < 50\%, N=15 (51.7\%) &	5-year mortality >= 50\% years, N=14 (48.3\%) &	p-value \\
     \hline
Age ± SD    & 69.07 ± 4.80	&66.86 ± 10.49	& 0.480 \\
\hline
Height ± SD	& 169.40 ± 4.07 &169.79 ± 11.23	& 0.905\\
\hline
Weight ± SD	& 83.53 ± 11.84 &84.07 ± 9.98	& 0.895\\
\hline
BMI ± SD	& 29.08 ± 3.74	&29.34 ± 3.96	& 0.858 \\
\hline
Sex (male)	& 15 (100.0)	& 11 (78.6)	& 0.100 \\
\hline
Heart Failure Duration (years) &5.08 ± 3.37 &8.80 ± 3.96	& \textbf{0.027} \\
\hline
EF ± SD	& 21.67 ± 9.19	&16.92 ± 9.47 &0.192 \\
\hline
Sodium ± SD	    &136.27 ± 3.97	&134.93 ± 4.83	&0.424 \\
\hline
BUN ± SD	    &36.71 ± 20.01	&40.50 ± 19.80	&0.612 \\
\hline
Creatinine ± SD	& 1.26 ± 0.40	&1.27 ± 0.53	& 0.967 \\
\hline
eGFR(mL/minute per 1.73 m2) ± SD & 61.53 ± 18.73 & 62.24 ± 30.16	& 0.940 \\
\hline
Uric Acid ± SD	& 7.85 ± 2.29	&7.61 ± 2.30  & 0.791 \\
\hline
NT-proBNP ± SD	& 8119.67 ± 8190.55	&19845.79 ± 15090.14 & \textbf{0.018} \\
\hline
BP systolic ± SD & 112.67 ± 14.78 & 182.57 ± 275.74	& 0.361 \\
\hline
Length of Hospitalization (Day) ± SD & 7.47 ± 1.85	& 6.62 ± 2.75	& 0.356 \\
\hline
Daily Furosemide Dose Requirement ± SD	& 256.00 ± 159.50	 & 387.69 ± 121.53	& \textbf{0.020} \\
\hline
HTN (\%)	    &15 (100.0)	&13 (92.9)	&0.466  \\
\hline
CAD (\%)	    &15 (100.0)	&6 (42.9)	&\textbf{<0.001}\\
\hline
CABG/PCI (\%)	&13 (86.7)	&6 (42.9)	&\textbf{0.021}\\
\hline
CKD (\%)	    &6 (40.0)	&7 (50.0)	&0.715\\
\hline
NYHA I   (\%)	&0 (0.0)	  &  0 (0.0)	& \\
\hline
NYHA II  (\%)	&2 (13.3)	 & 0 (0.0)	& \\
\hline
NYHA III (\%)	&13 (86.7)	& 5 (35.7)	&\\
\hline
NYHA IV  (\%)	&0 (0.0)	  & 9 (64.3) & \\
\hline
Chamber dilatation (\%) & 15 (100.0)  & 13 (92.9) & 0.333 \\
\hline
Left ventricular dilatation only (\%) &4 (26.7)	& 1 (7.1) & \\ \hline	
Left atrial and left ventricular dilatation (\%) &11 (73.3)	& 12 (85.7) & \\
\hline
Pretibial edema (\%)	&12 (80.0)	&13 (92.9)	&0.141 \\
\hline
Mild edema (\%)	        &3 (20.0)	&0 (0.0)	&      \\
\hline
Moderate edema (\%)	    &4 (26.7)	&2 (14.3)	&      \\
\hline
Severe edema (\%)	    &4 (26.7)	&9 (64.3)	&      \\
\hline
Very severe edema (\%)	&1 (6.7)	&2 (14.3)	&      \\
\hline
Pleural effusion (\%)	&3 (20.0)	&6 (42.9)	&0.245\\
\hline
\textbf{Acoustic Predictor} ± SD (CI) & 0.21 ± 0.19 & 0.81 ± 0.17 &\textbf{<0.001} \\
\hline
\multicolumn{4}{| p{15.0cm} |}{Abbreviations: MLE-mean life expectancy, SD-standard deviation, CI-confidence interval,  BMI-body mass index, EF-ejection fraction, BUN-blood urea nitrogen, eGFR-estimated glomerular filtration rate, NT-proBNP-N-terminal pro-b-type natriuretic peptide, BP-blood pressure, HTN-hypertension, CAD-coronary artery disease, CABG/PCI- coronary artery bypass grafting/percutaneous coronary intervention, CKD-chronic kidney disease, NYHA-New York Heart Association.} \\
[10ex]\hline
    \end{tabular}
\end{table*}

\subsection{Machine Learning Results} \label{sec: ml_results}
In this subsection, the results of our predictive model for assessing survival risk in heart failure (HF) patients are presented. Initially, an overview of the performance achieved by the machine learning model trained on the dataset is provided. Following that, a leave-one-out cross-validation (LOOCV) technique as a validation approach is employed to assess the model's ability to generalize to unseen data. Subsequently, building upon the results of the statistical analysis outlined in Table \ref{Table: patients}, the levels of one of the diagnostic biomarkers, NT-proBNP, are added to the dataset. The algorithm is then retrained to observe if this biomarker enhanced the model's classification ability as this hormone showcased a significant difference between the two classes in our statistical analysis (Table \ref{Table: patients})

\subsubsection{Evaluation of Logistic Regression on Train-Test Split}
As previously mentioned, the dataset was separated into training (N = 18 patients) and test (N = 11 patients) sets with a test ratio of 0.35. We focused on key evaluation metrics, including accuracy, precision, recall, and F1-score. The algorithm demonstrated performance metrics in both the training and test sets, as detailed in Table \ref{Table: results_train}. 

\begin{table}[h]
\centering
    \caption{\label{Table: results_train} Results of LR on Train-Test Split}
    \begin{tabular}{| p{ 1.8 cm} | p{1.1 cm} | p{1.1cm} | p{1.0cm}| p{1.2cm}|}
    \hline
    	& Accuracy & Precision	& Recall & F1-Score \\
     \hline
Training Set & 89\% & 1.0 & 0.78 & 0.875 \\
\hline
Test Set & 82\% & 0.80 & 0.80 & 0.80 \\
\hline
    \end{tabular}
\end{table}

\begin{figure*}[h] 
	\centering
 
\begin{subfigure}{}
    \includegraphics[width=0.4\linewidth,height= 0.31\linewidth]{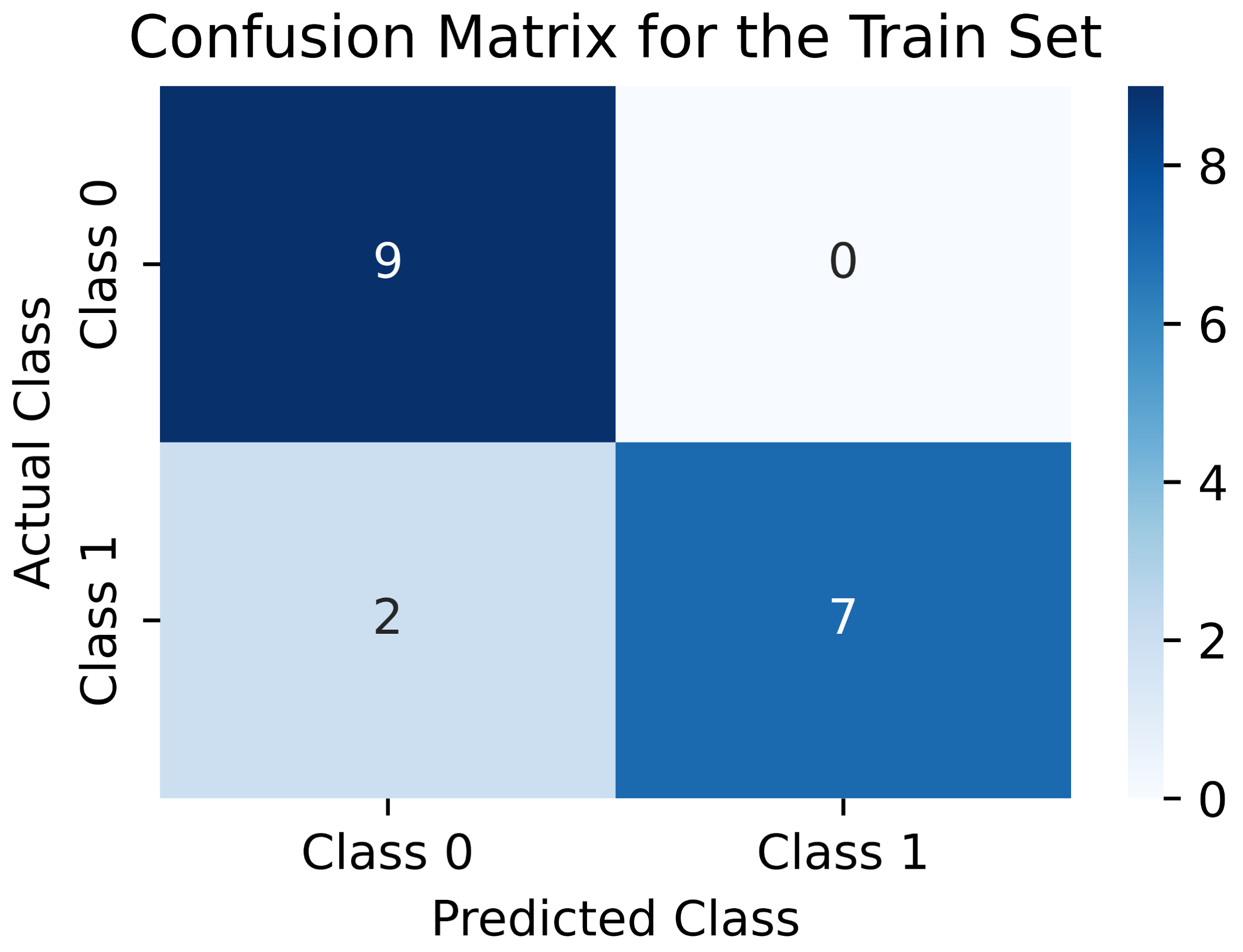}
\end{subfigure}
    \hfill
\begin{subfigure}{}
    \includegraphics[width=0.4\linewidth,height=0.31\linewidth]{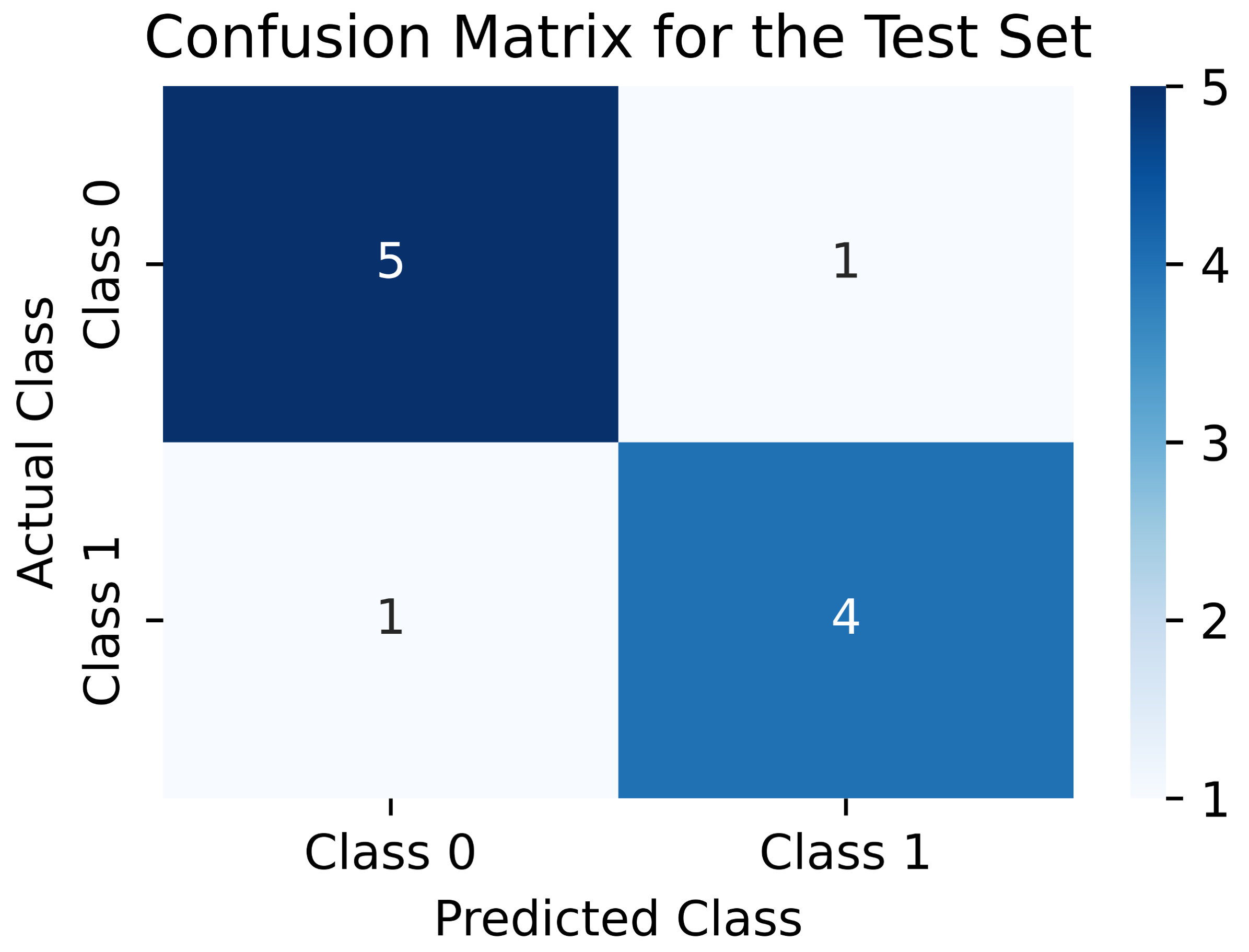}
\end{subfigure}
    
	\caption{\label{fig: conf_mat1} Confusion matrix evaluated on LR performance using hold-out dataset}
\end{figure*}

The confusion matrix evaluated on the developed logistic regression algorithm is presented in Fig. \ref{fig: conf_mat1}.
The Logistic Regression classification model was replaced by two additional conventional machine learning methods, namely Decision Tree (DT) and Random Forest (RF). After tuning the parameters in the preprocessing steps using the same parameter search space as the LR model, both algorithms achieved classification accuracies of 83\% and 82\% on the training and test sets, respectively.

\subsubsection{Evaluation of The Logistic Regression Model with Leave-One-Out Method}

\begin{table}[h]
\centering
    \caption{\label{Table: results_loocv} Results of LR with Leave-Out-Out method}
    \begin{tabular}{| p{ 1.8 cm} | p{1.1 cm} | p{1.1cm} | p{1.0cm}| p{1.2cm}|}
    \hline
    	& Accuracy & Precision	& Recall & F1-Score \\
     \hline
Training Set & 89.3\% & 0.923 & 0.86	& 0.89 \\
\hline
Test Set & 82.76\%	& 0.80	& 0.857	& 0.8276 \\
\hline
    \end{tabular}
\end{table}

To assess how well the developed model can generalize, a leave-one-out cross-validation (LOOCV) approach on the entire dataset was utilized. In this process, we cycled through all 29 samples, excluding one sample each time from the training set, while the model was trained on the remaining data. In this process, we cycled through all 29 samples, excluding one sample each time from the training set, while the model was trained on the preprocessed remaining data using the same preprocessing pipeline as in the train-test split method after the necessary fine-tuning of preprocessing and model hyperparameters. This effectively represents a split ratio of 28 for training and 1 for testing, resulting in a total of 29 unique combinations of training and test sets. Following the calculation of individual evaluation metrics of the model on 29 different training and test sets, overall metric values were obtained by averaging the evaluation metrics of each combination over training and test sets separately. Overall, the accuracy, precision, recall, and F1-score metrics in the training and test sets averaged over 29 combinations are shown in Table \ref{Table: results_loocv}.

\begin{figure}[h]
	\centering
		\includegraphics[width=1.0\linewidth, height=0.75\linewidth]{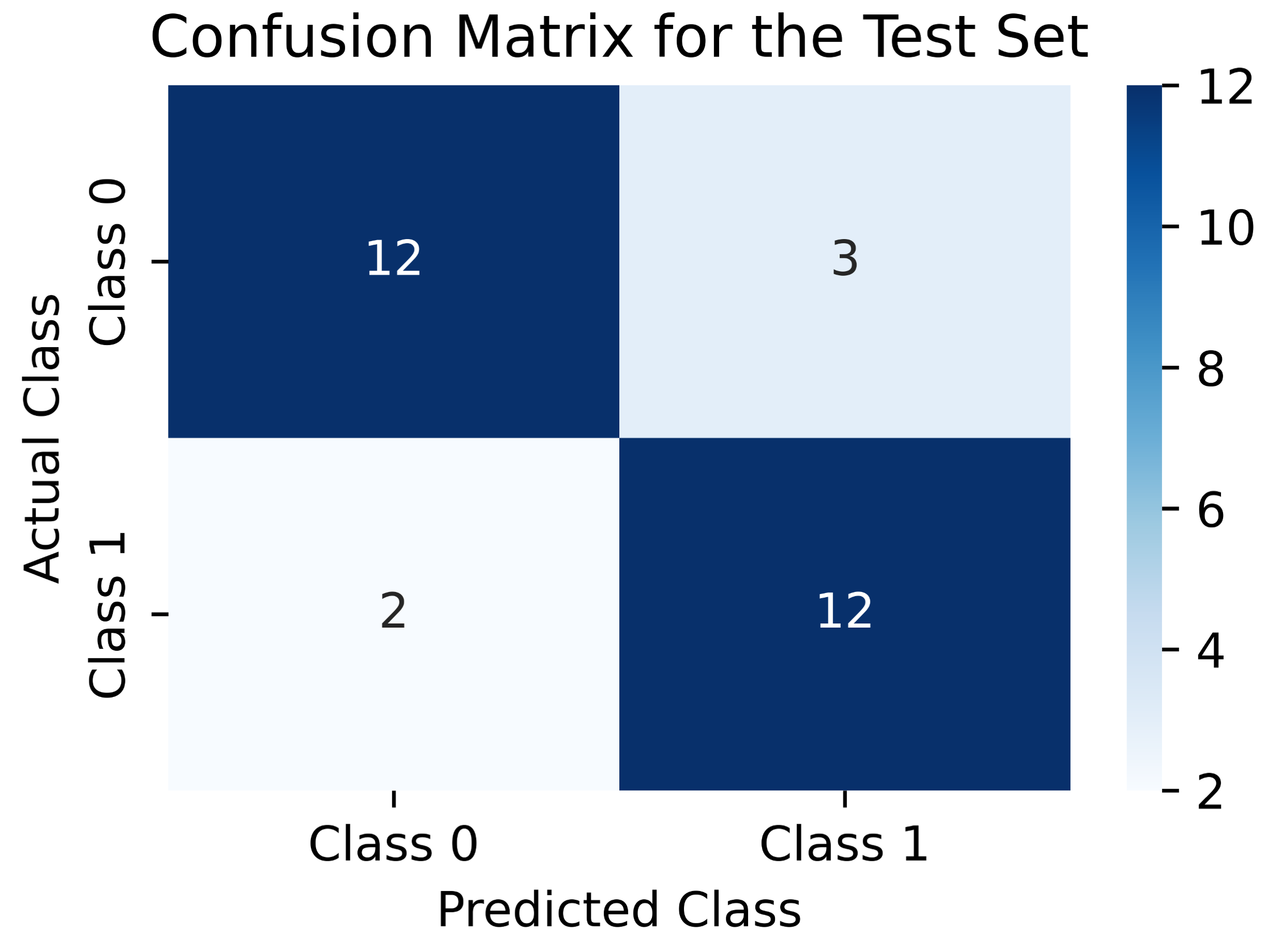}
	  \caption{Confusion matrix evaluated on LR with Cross-Validation}\label{fig: conf_mat2}
\end{figure}

The corresponding confusion matrix formed as a result of test set predictions is presented in Fig. \ref{fig: conf_mat2}.

These findings highlight the model's robustness and its capability to effectively generalize across a range of test scenarios, making it a promising tool for the targeted application. Furthermore, as shown in Table \ref{Table: patients}, NT-proBNP, a widely employed diagnostic biomarker in clinical medicine as an indicator for cardiac dysfunction, shows a statistically significant difference between the 2 classes. Based on this observation, the model was retrained by incorporating this hormone’s in-blood level as an additional feature to the dataset. The model’s performance was then evaluated to assess the additional information provided by the new feature. Repeating the Leave-One-Out technique with identical parameter settings of the developed model, higher accuracies of 96.4\% and 86.2\% were achieved this time, compared to 89.3\% and 82.76\% with acoustic predictors alone. This showcases the relevance of NT-proBNP in examining the mortality rate of HF patients.

\subsection{Comparative Analysis with Existing Works}
Our model's performance is comparable to, and even slightly better than, existing models in terms of predictive accuracy, sensitivity, and specificity. For example, the deep learning system developed by \cite{ml_mortality4} using medical data from the MIMIC-III v1.4 dataset achieved a precision of 0.76, a recall of 0.81, and an F1 score of 0.78. In comparison, our model demonstrates better performance with a precision of 0.80, an F1 score of 0.83, and a recall of 0.86. The study by \cite{ml_mortality5} evaluates the performance of various machine learning algorithms in predicting outcomes for heart failure patients. The study employed several supervised machine learning algorithms, including logistic regression (LR), decision tree (DT), and random forest (RF). The highest accuracy reported was 80\%, which is slightly lower than our accuracy of 82.8\%. Moreover, their study showed a very low specificity value (0.11), compared to our 0.80, indicating its inefficacy in correctly identifying negative cases.

\section{Discussion and Future Work} \label{sec: discuss}
In this paper, the potential of voice analysis as a non-invasive predictive tool for Heart Failure (HF) and the role of Artificial Intelligence (AI) in advancing HF diagnosis was showcased. Our study aimed to fill a gap in the literature by exploring the link between voice changes and mortality in HF patients, utilizing standardized speech protocols for voice analysis among HF patients. Comprehensive data linking voice characteristics to disease conditions and their effects on patient health are currently insufficient. While various studies have hinted at potential links between voice patterns and conditions such as attention deficit hyperactivity disorder (ADHD), Parkinson's disease, and dyslexia \cite{parkinsons, adhd, neuro}, research specifically examining the relationship between voice and cardiovascular disease is notably limited. 
Recent studies have integrated automated speech analysis technology to directly detect specific complications, such as pulmonary fluid overload, during hospitalization, and have analyzed changes in voice features over time to understand the effectiveness of treatment for acute decompensated heart failure (ADHF) \cite{murton2023acoustic}. Notably, \cite{maor2020} came up with a noninvasive vocal biomarker that is associated with adverse outcomes among patients with Congestive Heart Failure (CHF) through thorough statistical analysis. Their study has explored the correlation, but it has not developed a model to predict the mortality rate solely using patient vocal features.

To the best of our knowledge, this study represents the first attempt to employ a machine learning-based method for assessing the mortality rate among acute decompensated heart failure patients based on acoustic changes. We deploy a fine-tuned logistic regression model backed by a heuristically tailored pre-processing and feature selection phase for training and testing the collected samples. Subsequently, reproducibility is examined through a cross-validation approach to assess the model's generalization ability to unseen data. The model demonstrates consistent results, with a slight increase in accuracy from 82\% to 82.76\% as a result of cross-validation. Moreover, the acoustic predictor generated by our machine learning model shows a significant correlation with patient labels (p=<0.001), underscoring the statistical significance of the relationship between the predictor and patient outcomes. Incorporating the levels of the statistically significant biomarker NT-proBNP into the dataset further improves the model, leading to training and test set accuracies of 94.4\% and 91\%, respectively, using the train-test split methodology. Additionally, an enhancement in accuracies from 82.76\% to 86.2\% is observed with the cross-validation approach. This underscores the significance of NT-proBNP in evaluating the mortality rate of HF patients. Furthermore, two additional ML classifiers, decision tree, and random forest, were also utilized after undergoing the same preprocessing pipeline for hyperparameter tuning. While their outcomes are notably strong, logistic regression modeling demonstrated a better discriminatory power compared to these alternatives.
An advantage of our study over the SHFM model lies in the simplicity of our approach, requiring only the patient's voice as input. In contrast, SHFM relies on intricate medical parameters provided by a qualified medical professional, relying heavily on comprehensive clinical data and blood samples.

Our methodology diverged from conventional approaches such as LASSO and Elastic Net for feature selection due to the impracticality of employing them in scenarios where the number of features outnumbers the sample size. Such an impracticality is saturation in LASSO’s variable selection process. LASSO’s constraints, which necessitate the number of selected variables to be smaller or equal to the sample size, often result in the selection of only a single feature from highly correlated sets. Despite Elastic Net’s capability to handle feature selection beyond the sample size, its efficacy diminishes in cases where inadequate sample sizes impede effective subsampling, resulting in suboptimal performance \cite{lassobackward3, Hastie2001, elasticnet, lasso-elastic1, lasso-elastic2, lasso-elastic3}.

Interpreting the chosen features derived from the feature engineering and modeling methodology applied during the train-test split, as illustrated in Figure {\ref{fig:boxplot}}, it is evident that the selected features for effective patient classification encompass all sections of the voice recordings and vocal feature categories delineated in Sections {\ref{sec: capev}} and {\ref{sect: FeatureExtraction}}. Specifically, patients with a mortality rate exceeding 50\% exhibit notably higher global harmonic richness factors and lower spectral tilt skewness parameters obtained in sections 1 and 3, of audios with controlled protocol, implying an increase in harmonic speech components above the fundamental frequency and indicating a weaker harmonic structure in the glottal source spectrum. This observation aligns with findings from \cite{MITTAPALLE202235}, where a similar assessment of harmonic richness revealed analogous trends. Furthermore, patients with mortality rates surpassing 50\% demonstrate increased statistical standard deviations in the mean squared error between original and reconstructed F0 values in section 1, of audios with controlled protocol, indicative of heightened potential irregularities in vocal pitch control. Additionally, elevated standard deviations in amplitude perturbation quotient (apq) values across vocal frames derived from section 4 suggest greater variability in vocal fold vibratory patterns or phonatory behavior inconsistencies in responses to directed questions and spontaneous speech. Additionally, elevated standard deviations in amplitude perturbation quotient (apq) values across vocal frames obtained from section 4 suggest greater variability in vocal fold vibratory patterns or phonatory behavior inconsistencies in responses to directed questions and spontaneous speech, aspects often examined in the context of Parkinson's disease for implications on monotonicity and intelligibility \cite{VASQUEZCORREA201821}. Although the APQ parameter, which reflects the cords' inability to support periodic vibration and reveals hoarse and breathy voices \cite{Liaw2020}, exhibits distinct mean values across section 2 speech segments for both groups, those with a higher mortality rate demonstrate greater variability compared to the other group.

Assessing NT-proBNP's impact as an input parameter, the model incorporating NT-proBNP yielded a coefficient of 0.39 for equation {\ref{eq: z}} and an odds ratio of 1.48. Notably, this coefficient is assigned to the scaled version of the parameter's true values. Upon reversing the transformation to revert to the original values, it is revealed that for every 9800-unit increase in the NT-proBNP predictor variable, the likelihood of a patient being predicted to have a 5-year mortality rate exceeding 50\% increases by almost 50\%.

Another advantage of our study is that, intriguingly, the age distribution in Table {\ref{Table: patients}} reveals that patients at higher risk of mortality have a mean age lower than those at lower risk, yet the averages of both groups are sufficiently close and comparable. This ensures robust and reliable modeling, preventing feature selection bias towards age-related features. Likewise, the p-value of patient age for the target label was found to be statistically insignificant.

However, a limitation of our study is the gender distribution imbalance among the samples, with a majority of 26 out of 29 patients being male.
Although all three female patients in the dataset were assigned the same target label, suggesting a potential gender bias, examination of Figure {\ref{fig:boxplot}} and Section {\ref{sec:featureExplanation}} reveals that the features utilized may not be directly related to gender, particularly those related to pitch frequency. However, we observed that the misclassified samples consistently belonged to this minority gender group. These misclassified female patients also exhibited values of selected features that were closest to the opposite target label, indicating outliership within their assigned targets. This phenomenon may be attributed to variations in physiological vocal tract formations influencing the selected features. Interestingly, in cases where NT-proBNP was included in the dataset, misclassified samples had NT-proBNP values closest to the mean values of the opposite class.

Furthermore, the sample size used in this study is relatively small when compared to the extensive body of machine learning literature. While small sample sizes are common and characteristic in biomedical and biological research, numerous studies highlight the challenges in balancing and satisfying the bias-variance trade-off in such scenarios \cite{cohenHudson, SHAIKHINA2015469, Lever2016}.

Nevertheless, similar studies suggest using the leave-one-out cross-validation technique to assess the position in the bias-variance space of the model. Setting the number of splits (K) close to the number of samples (N) further reduces the likelihood that a split will result in sets that are not representative of the full dataset, thus minimizing underfitting or overfitting. Furthermore, the preprocessing and feature selection phase, which is heuristically tailored and detailed in Section {\ref{sec: feature engineering}}, tackles this issue by eliminating features with sporadically high MI scores from randomly formed subsets.

Nonetheless, both the preprocessing steps and the classification model construction have achieved remarkable accuracy.

Addressing this imbalance and incorporating a larger dataset with a more even gender distribution would contribute to the development of a more robust model that accounts for gender-related differences in vocal features. Additionally, our model's reliance on supervised training, considering the SHFM model results as ground truth, poses a limitation, as SHFM was specifically designed and verified using data from outpatient participants involved in clinical trials, observational studies, or clinical registries. Therefore, its suitability may be limited, especially for those hospitalized or dealing with significant health conditions beyond heart failure. Furthermore, there is a need for ongoing research to explore the integration of vocal biomarkers into the daily monitoring routine of patients. Continuous updates and adaptations are crucial to keep pace with the evolving landscape of heart failure management and advancements in medical knowledge.

Further exploration into the integration of additional non-invasive biomarkers, such as wearable device data (e.g., heart rate, physical activity), could improve the predictive power of the model. Combining voice analysis with these other data sources may yield a more comprehensive assessment of patient health. Developing real-time monitoring systems that utilize our predictive model to continuously assess patient risk and provide early warnings to healthcare providers could facilitate timely interventions and potentially improve patient outcomes.

\section{Conclusion}
In conclusion, our study highlights the potential of voice analysis and artificial intelligence as a non-invasive means to predict mortality rates in heart failure (HF) patients. The logistic regression model outputs an Acoustic Predictor, a combination of important vocal features that shows a strong correlation with the 5-year mortality rate (p < 0.001). The model demonstrated robust predictive capabilities, providing a user-friendly alternative to traditional clinical parameters. Notably, the incorporation of the NT-proBNP biomarker significantly enhanced the model's performance. Recognizing minor constraints in areas like sample size and gender distribution, our study lays a foundation for future research endeavors focused on broadening datasets and seamlessly incorporating vocal biomarkers into routine patient monitoring. This innovative approach holds significant potential to improve patient outcomes, streamline resource allocation, and advance patient-centric heart failure management. It leads to a future where diagnosing a patient's condition through a simple phone speech not only reflects the promise of widespread accessibility and convenience but also represents a transformative leap forward in healthcare innovation.

\printcredits



\bibliographystyle{elsarticle-num} 
\bibliography{main}



\end{document}